\documentclass[10pt]{article}

\usepackage{amsmath, amsfonts, amssymb, mathrsfs}
\usepackage{lineno}
\usepackage{hyperref}

\modulolinenumbers[5]

\usepackage{wrapfig}

\def\drawplusplus#1#2#3{\hbox to 0pt{\hbox to #1{\hfill\vrule height #3 depth
      0pt width #2\hfill\vrule height #3 depth 0pt width #2\hfill
      }}\vbox to #3{\vfill\hrule height #2 depth 0pt width
      #1 \vfill}}
\def\concat{\mathrel{\drawplusplus {12pt}{0.4pt}{5pt}}}

\usepackage{amsmath, amssymb, amsthm}

\usepackage[all]{xy}
\usepackage{graphicx}
\newtheorem{thm}{Theorem}

\begin{document}


\title{Folding and Unfolding on Metagraphs}

\author{Ben Goertzel}





\maketitle

\begin{abstract}
Typed metagraphs are defined as hypergraphs with types assigned to hyperedges and their targets, and the potential to have targets of hyperedges connect to whole links as well as targets.   Directed typed metagraphs (DTMGs) are introduced via partitioning the targets of each edge in a typed metagraph into input, output and lateral sets; one can then look at "metapaths" in which edges' output-sets are linked to other edges' input-sets.    An initial algebra approach to DTMGs is presented, including introduction of constructors for building up DTMGs and laws regarding relationships among multiple ways of using these constructors.   A menagerie of useful morphism types is then defined on DTMGs (catamorphisms, anamorphisms,  histomorphisms, futumorphisms, hylomorphisms, chronomorphisms, metamorphisms and metachronomorphisms), providing a general abstract framework for formulating a broad variety of metagraph operations.  Deterministic and stochastic processes on typed metagraphs are represented in terms of forests of DTMGs defined over a common TMG, where the various morphisms can be straightforwardly extended to these forests.   The framework outlined can be applied to realistic metagraphs involving complexities like dependent and probabilistic types, multidimensional values and dynamic processing including insertion and deletion of edges. 
\end{abstract}



\tableofcontents

\section{Introduction}

In this paper we take a few steps toward bringing together two directions of formal and conceptual development that have advanced significantly in recent years, and that we believe can be even more impactful in the coming phase if they are appropriately allied together:

\begin{itemize}
\item The use of generalized graph-like models to represent knowledge -- e.g. hypergraphs (graphs where links can span more than two nodes) and metagraphs \cite{basu2007metagraphs} (which include also links pointing to links, and links pointing to whole subgraphs)
\item  The use of fold and unfold (catamorphism, anamorphism and related) operations to abstractly represent a variety of algorithmic operations
\end{itemize}

Most of the work with fold-based representation of algorithms has centered on folding over lists and trees,but there have also been extensions of the core concepts to a broader scopes of data structures \cite{erwig1998metamorphic}.    Gibbons \cite{gibbons1995initial} has provided a simple and elegant formalization of folding over directed multigraphs.

Here, inspired by Gibbons' \cite{gibbons1995initial} work and extending beyond it in significant ways, we propose a natural way to define folding, unfolding and related operations over directed, typed metagraphs.

We show that the ideas outlined can be applied to realistic metagraphs involving complexities like dependent and probabilistic types, multidimensional values and dynamic processing including insertion and deletion of edges.   And we explain how  the various morphisms defined can be straightforwardly extended to forests of DTMGs defined over a common TMG, a structure that emerges naturally when one tries to represent families of deterministic or stochastic processes over TMGs.

\subsection{Background}

The purpose of this work will be clearest to the reader who is familiar with the substantial  literature descended from Meijer, Fokkinga, and Paterson's classic 1991 paper {\it Functional programming with bananas, lenses, envelopes and barbed wire} \cite{meijer1991functional} and related works published around that time, exploring the use of hylomorphisms (compositions of catamorphisms with anamorphisms) to represent various sorts of algorithms.  This line of research has since blossomed in multiple directions, including the use of Galois connections to provide abstract derivations of algorithms falling into broad categories such as greedy or dynamic-programming-like  \cite{mu2012programming}, and the identification of various more refined morphism types building on the basic catamorphism, anamorphism and hylomorphism \cite{bracht2016}.   Extensions like histomorphisms and futumorphisms, which persist long-term memory through the folding and unfolding process, extend the scope of algorithms that can be cleanly represented in this sort of framework.

The use of hypergraphs and metagraphs for algorithmics of various sorts is a broad topic which is not restricted to one subfield of AI or computer science.   The specific motivation for the current research has largely been the application of these structures in Artificial General Intelligence system design, as outlined e.g. in \cite{EGI1} \cite{EGI2}.   The  practical computational advantages of hypergraphs and metagraphs over straightforward graphs in this context have been eloquently reviewed by Linas Vepstas in \cite{VepstasRAM2020}.   However the application of the formalizations presented in this paper are likely significantly. more broad than this particular thread of work applying metagraphs in an AGI context.

\section{Directed Typed Metagraphs }

The intuitive notion of a metagraph is quite simple -- nodes and links, where links can point to nodes, links or whole subgraphs, and nodes and links can be labeled or weighted as one wishes \cite{basu2007metagraphs}.   However, formalizing the notion in a way that suits both practical applications and theoretical analysis requires a bit of care.

\subsection{Typed Metagraphs}

We are concerned here with "typed metagraphs" defined as follows.   

Let $\mathcal{T}$ be a space of type expressions, which comes with a notion of type inheritance, and a root type $e$ from which every type inherits.   We will say $T_1 \preccurlyeq T_2$ if $T_1$ inherits from $T_2$, and $\mathcal{C}(T_1, T_2)$ if $T_1$ and $T_2$ are comparable, i.e. if $T_1 \preccurlyeq T_2$  or $T_2 \preccurlyeq T_1$ .

Let $[n] = \{1,\ldots,n\}$.   An {\it edge}  of arity $n$ is characterized by a {\it typed-target tuple} $\mathcal{L}$ of the form

$$
\mathcal{L} = \left( T, (L_1, T_1 )  , \ldots , (L_n, T_n )  \right)
$$

\noindent where $T_i \in \mathcal{T}$ and $L_i \in [n]$.    

\begin{figure}
  \includegraphics[width=10cm]{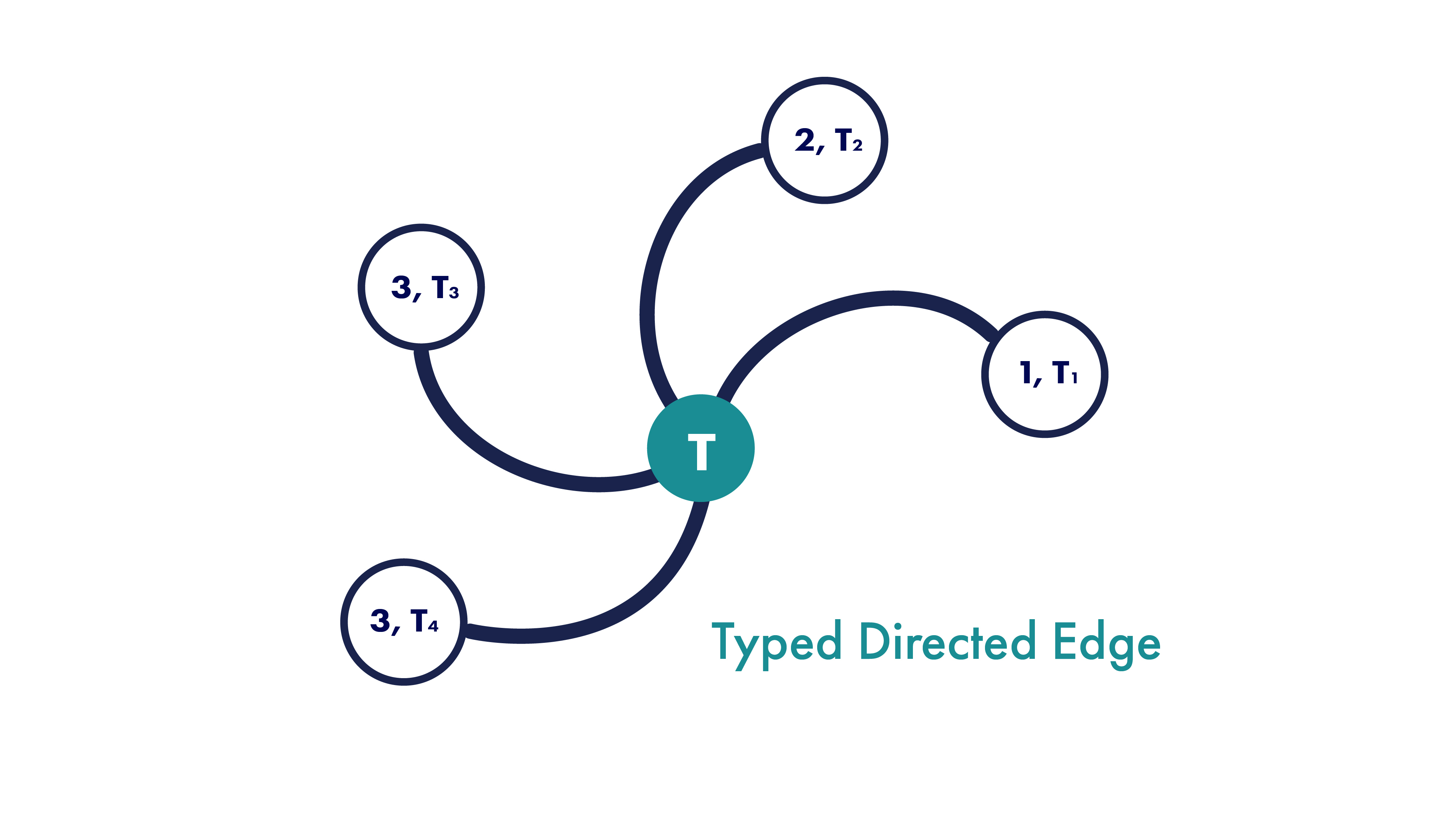}
  \caption{Example of a typed, directed edge.   This one has $4$ targets.  $T$ is the type of the edge itself.   The third and fourth targets are unordered relative to each other.}
  \label{fig:Edge}
\end{figure}

In this definition, the initial $T$ denotes the type of the link as a whole.   The following $n$ ordered pairs correspond to the $n$ targets of the edge.  The $i$'th target is characterized by the type $T_i$ and an an index label $L_i$ that determines where the target lies in the ordering of targets.   Multiple targets may have the same index label; if all targets have the same index label then the edge is effectively unordered.    

These "edges" are similar to what Linas Vepstas called "seeds" or "germs" in \cite{vepstas2019sheaves}.

For an edge $E_i$ we will write

$$
\left( T^i, (\mathcal{L}^i_1, T^i_1 )  , \ldots , (\mathcal{L}^i_n, T^i_n )  \right)
$$

\noindent A {\it typed metagraph}  or {\bf TMG} is then defined as a pair $G =(E, C)$ where $E$ is a set of edges, and  $C$ is a set of tuples of the form $(E_{r1}, E_{r2},t_{r1} , t_{r2})$ where $E_{r1}$ and $E_{r2}$ are edges in $E$, and $t_{r1}$ and  $t_{r2}$ are targets of $E_{r1}$ and $E_{r2}$ respectively.    These tuples may be thought of as {\it connections} joining edges.   The numbers $t_{r1}$ and $t_{r2}$ indicate the index labels of the targets of the edges of $E_{r1}$ and $E_{r2}$ involved, where these number are set to $0$ if the target is the whole link.

We write ${\mathcal M}$ for the space of typed metagraphs.

More formally, where $E_{r1}$ and $E_{r2}$ are edges of arity $n_{r1}, n_{r2}$ respectively: If $(E_{r1}, E_{r2},t_{r1} , t_{r2}) \in C$ then this implies that $t_{r1} \in [ n_{r2}]$ and $t_{r2} \in [ n_{r2}]$, and one of the two conditions holds:

\begin{enumerate}
\item $t_{r1} \neq 0$, $t_{r2} \neq 0$, and $\mathcal{C}(T^1_{t_{r1}}, T^2_{t_{r2}})$
\item Either  
\begin{itemize}
\item $t_{r1}=0$ ,  $t_{r2} \neq 0$, and $\mathcal{C}(T^1, T^2_{t_{r2}})$
\item$t_{r2}=0$ ,  $t_{r1} \neq 0$, and $\mathcal{C}(T^1_{t_{r2}}, T^2)$
\end{itemize}
\end{enumerate}

\noindent In the first case, the interpretation is that the ${t_{r1}}$'th target of the edge $E_{r1}$ is connected to the ${t_{r2}}$'th target of the edge $E_{r2}$.    In the second case, e.g. if $t_{r2}=0 $ then the interpretation is that the ${t_{r1}}$'th target of the edge $E_{r1}$ is connected to the edge $E_{r2}$ as a whole (rather than to any of its targets).

In this framework, a metagraph may have dangling targets, i.e. targets of some of its edges that are not connected to anything.   The dangling targets will be labeled with types.    The set of dangling targets of a metagraph itself has the form of an edge (it's a collection of targets, each of which comes with a type and an index label).

To conceptually represent an edge pointing to a whole metagraph, one can create an unordered edge with a target connected to each edge in the metagraph, and assign a special type $M$ to this edge (a type connoting "metagraph wrapper").

\subsubsection{Probabilistically Typed Metagraphs}

A natural extension is to the case where the types on edges and targets are probabilistic dependent types, e.g. according to the PDTL formalism developed by Warrell \cite{warrell2016}.   In this case we have probabilistic type inheritance relations $T_1 \preccurlyeq_\rho T_2$  and $\mathcal{C}_\rho(T_1, T_2)$, and based on these we can assign a probability $\rho$ (or a whole probability distribution) to each connection between edges.   The evaluation of these probabilities is naturally done via sampling, or via probabilistic inference methods that more inexpensively approximate the results of such sampling under appropriate assumptions.

\subsubsection{Valued Metagraphs}

For many applications, it is also interesting to associate metagraph edges $E_i$ with finite lists $\mathcal{V}_i$ of {\it values}, each of which may be integer, floating-point or more complex in structure.   Such values are largely orthogonal to the core considerations in this paper, but are important to consider as they are critical to so many practical metagraph utilizations.   In the following we will assume that the TMGs involved may be valued, without giving much comment to the matter as it's mostly not material to the matters at hand.

\subsection{Directed Typed Metagraphs}

Suppose we have a TMG  $G$ whose dangling targets are divided into three categories: Inputs ($G_I$), Outputs ($G_O$) and Lateral ($G_L$).   We may then consider $G$ as an   {\it atomic directed typed metagraph} (atomic DTMG) which points from the inputs to the outputs.   (I.e. conceptually one is considering the whole $G$ as a single directed edge.)

\begin{figure}
  \includegraphics[width=10cm]{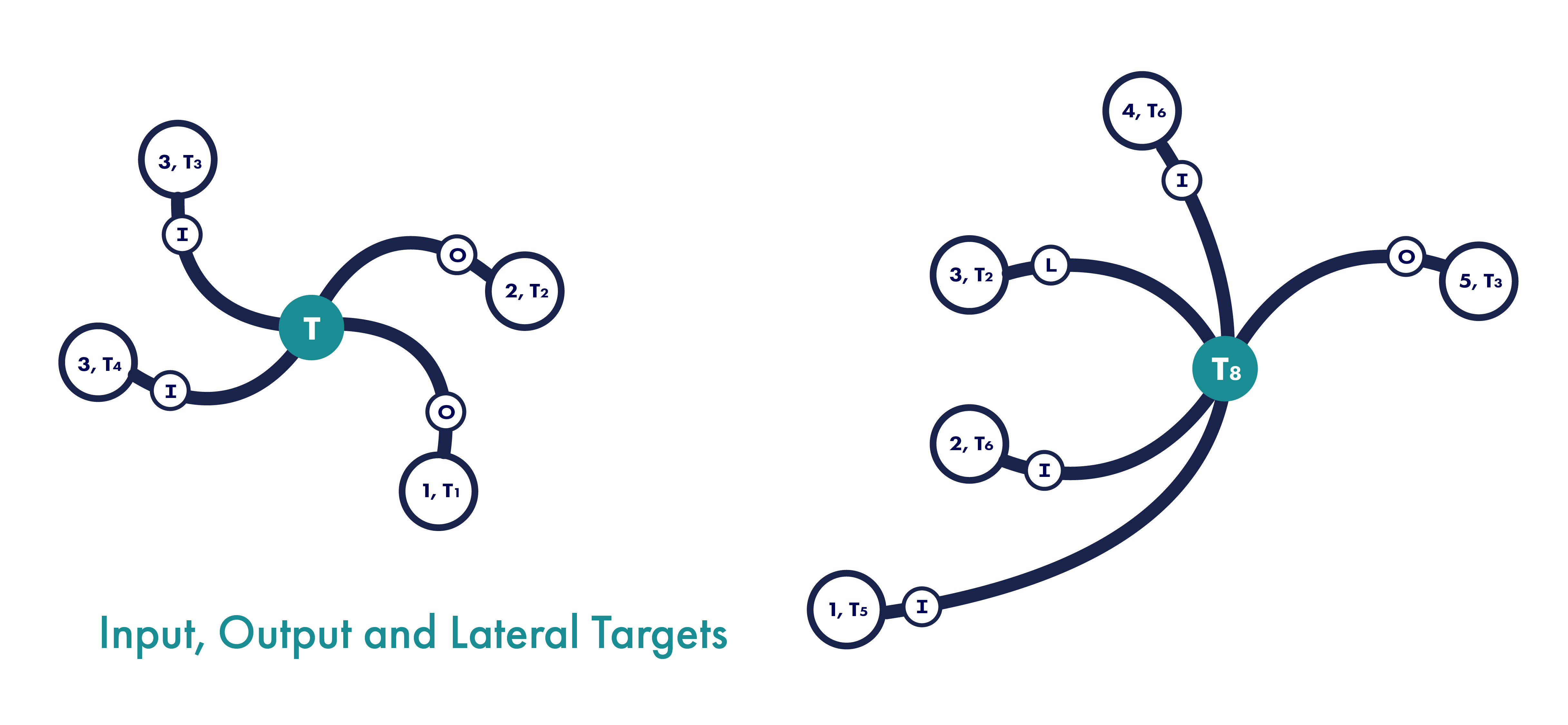}
  \caption{Partitioning of the targets of an edge into Input, Output and Lateral sets}
  \label{fig:Targets}
\end{figure}

A multigraph as considered in \cite{gibbons1995initial} is obtained if one restricts attention to binary links and then sets $G_I$ as the dangling targets with index $1$ and $G_O$ as the dangling targets with index $2$.

If $G$  is a subgraph \footnote{I will use the term "subgraph" instead of "submetagraph" just for simplicity and concision} of another TMG $G^1$, then  $G$ may be considered an atomic DTMG  as well.   In this case one needs to look at the dangling targets $G$ would have if it were extracted from $G^1$.  We may call these {\it external targets}, as distinct from {\it internal targets}, which are targets of edges contained in $G$ that are connected to targets of edges contained in $G$.  Dangling targets may be considered as a special case of external targets.   So we may then consider an atomic DTMG more generally as a TMG whose {\it external targets} (including but not restricted to its dangling targets) are divided into three categories: Inputs ($G_I$), Outputs ($G_O$) and Lateral ($G_L$).   

Suppose one has a metagraph $G^1$ in which the DTMG $G$ is a subgraph.   Then we say $G^2$ provides {\it input} to $G$ if every external target in $G_I$ is connected to some external target in $G^2_O$.    Similarly, we say $G^3$ receives {\it output} from $G$ if every external target in $G_O$ is connected to some external target in $G^3_I$. 

We may say that a DTMG $G$ forms a {\it metapath} from $G^2$  to $G^3$ if $G$ that receives input from  $G^2$  and provides output to  $G^3$.

\begin{figure}
  \includegraphics[width=10cm]{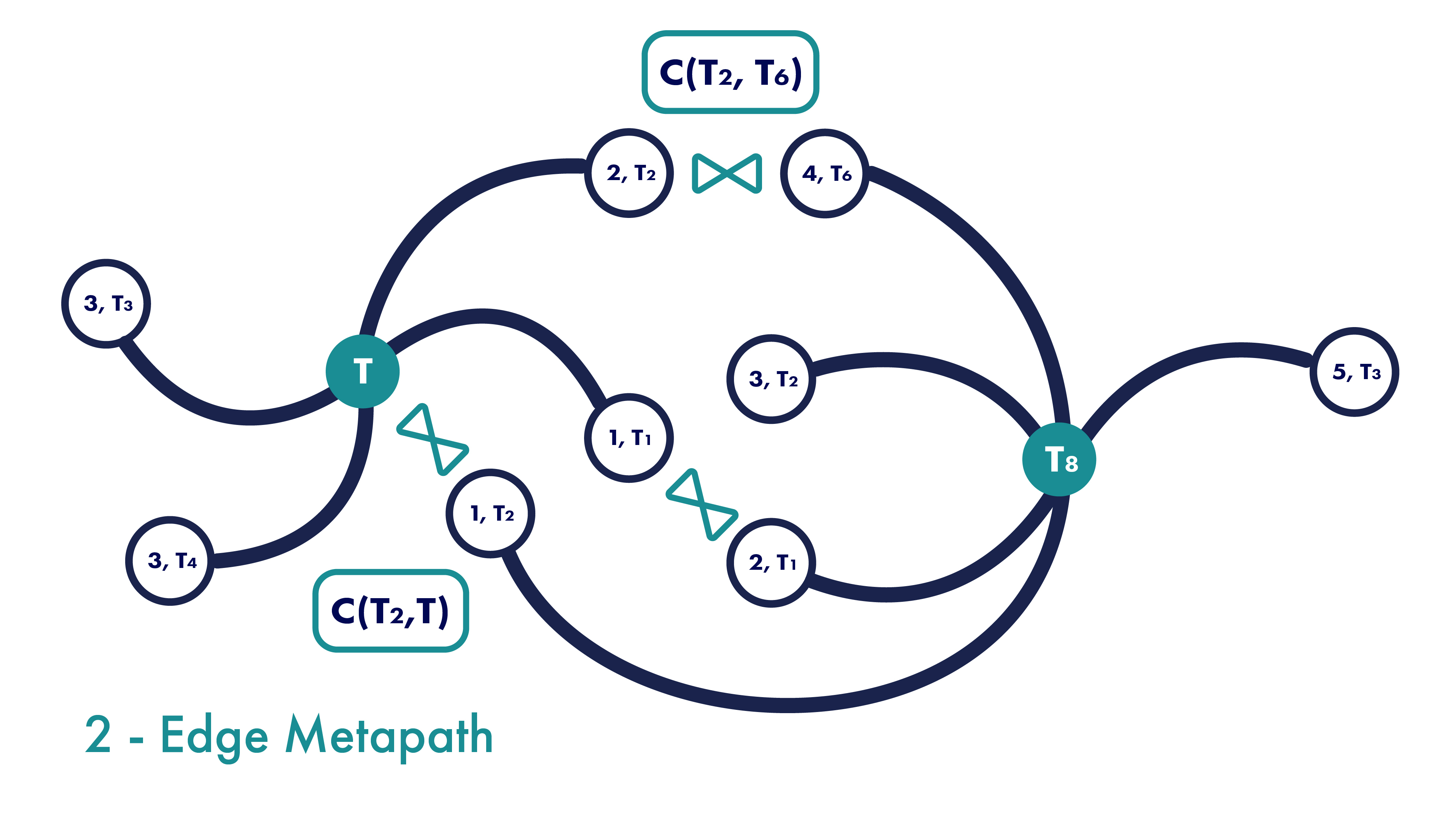}
  \caption{A short metapath, formed by connecting two directed edges.}
  \label{fig:Metapath}
\end{figure}

A DTMG may then be generally defined as {\bf a TMG composed by connecting DTMGs via metapaths} -- a recursive definition that bottoms out on the definition of atomic DTMGs.

We write ${\mathcal D}$ to denote the space of DTMGs.

(Note that a DTMG is a metapath and vice versa; a metapath is merely a DTMG viewed from the perspective of its origin and destination.)

In the extension to probabilistic types, each metapath naturally comes with a probabilistic weight, which gives the probability that there is type compatibility among the targets being connected.

\section{Constructing Directed Typed Metagraphs}

In this section we present an initial-algebra description of directed typed metagraphs.

\subsection{Constructors for Directed Typed Metagraphs}

We start  via presenting a set of operations for constructing DTMGs.  

\subsubsection{Edge}

The first DTMG constructor is the edge constructor itself, which builds an edge from a typed-target tuple and (if the metagraph is valued) a list of relevant values.

\subsubsection{Beside}

Given two DTMGs $G$ and $H$, the combination $G \Vert H$ ($G$ beside $H$) informally consists of the two DTMGs operating in parallel.   The input and output sets of the parallel combination are given by $(G \Vert H)_I = G_I \concat H_I$, $(G \Vert H)_O = G_O \concat H_O$ (where $\concat$ denotes concatenation).

The $\Vert$ operator evidently obeys an associative law, and the notation $m \times G$ may be used for the DTMG formed by placing $m$ copies of $G$ beside each other.

\begin{figure}
  \includegraphics[width=10cm]{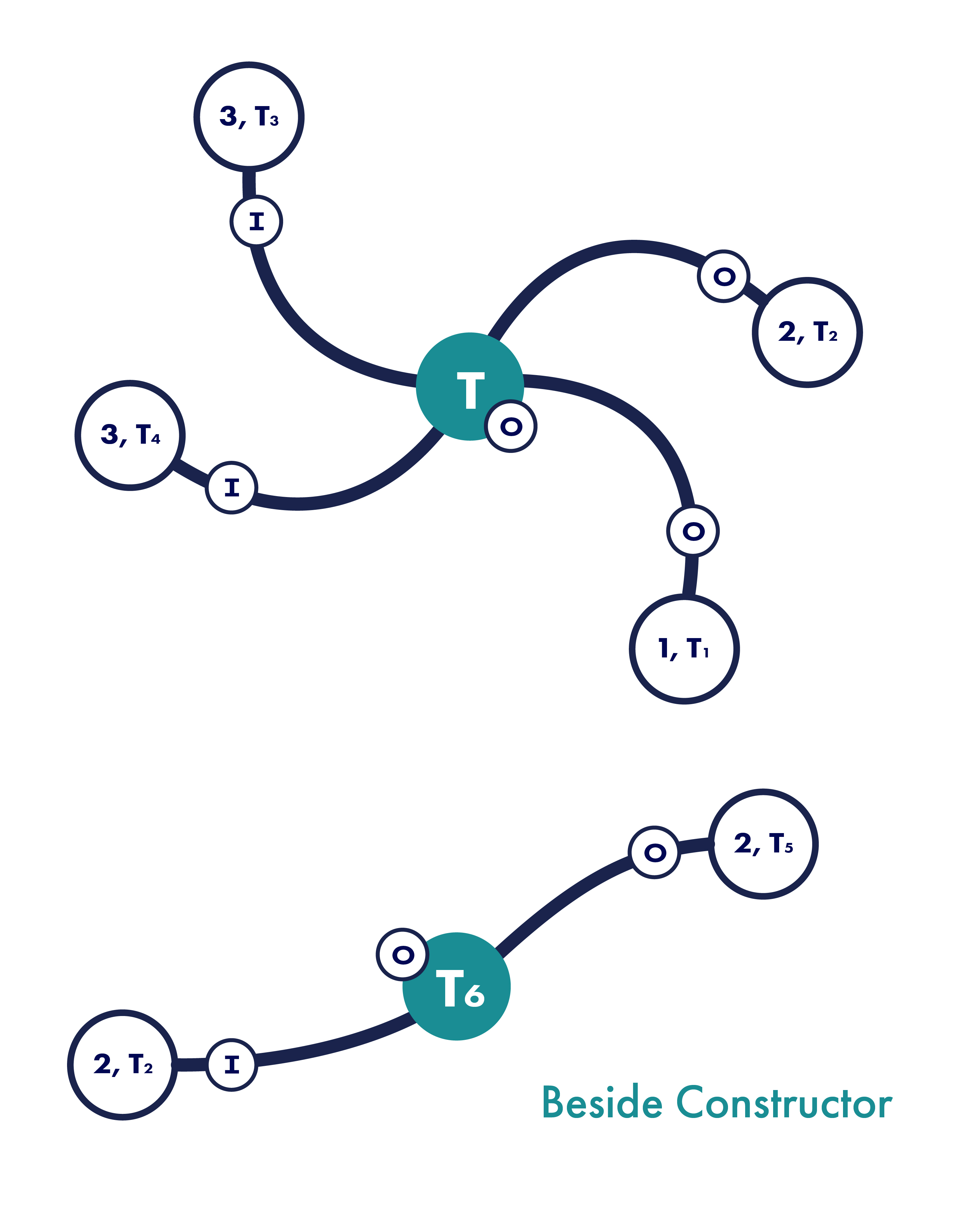}
  \caption{The Beside constructor.   The inputs [outputs] of the parallel combination of two DTMGs is the concatenation of their inputs [outputs].}
  \label{fig:Beside}
\end{figure}

\subsubsection{Connect}

A connection operation between two DTMGs $G$ and $H$ is one that connects some of $G$'s outputs to some of $H$'s inputs in a way that is compatible with the types of the targets involved (so an output target with type $\mathcal{T}_1$ gets matched to an input target with type $\mathcal{T}_2$ only if $\mathcal{C}(\mathcal{T}_1, \mathcal{T}_2)$, i.e. if the types are compatible).

If $G$ has $m$ outputs and $H$ has $m$ inputs, then there are at most 

$$
\sum_{i=1}^{min(m,n)} \frac{ max(m,n)! }{(max(m,n) - min(m,n)   )!  }
$$

\noindent possible connection operations between $G$ and $H$.   Each invertible {\it connection routing function} or {\it crf} 

$$
P: s^k_{[m]} \rightarrow s^k_{[n]}
$$

\noindent where $|s^k_{[m]}| = |s^k_{[n]}| = k, s^k_{[m]} \in 2^{[m]} , s^k_{[n]} \in 2^{[n]}$ potentially defines a connection operation, though not all of these will be type compatible.   The  parameter $k$ indicates the ${\it size}$ of $P$, meaning the cardinality of $s^k_{[m]}$ (which is also the cardinality of $s^k_{[n]}$), i.e. the number of linkages between targets that $P$ makes.

We will write $G \bowtie_P H$ to indicate connection of G to H according to the crf $P$.

The set of $\bowtie_P$ operators for various $P$ are mutually associative.   Furthermore, a version of the "abiding law" from \cite{bird1989lectures} holds as follows:

$$
(G \bowtie_P H) \Vert (J \bowtie_Q K)  = (G \Vert J) \bowtie_{P \Vert Q} (H \Vert K)
$$

\noindent where $P \Vert Q$ is a connection operation that maps $m_P+m_Q$ outputs into $n_P + n_Q$ inputs, via mapping its first $m_P$ arguments into its first $n_P$ output slots and then mapping its remaining $m_Q$ arguments into the remaining $n_Q$ output slots.

\begin{figure}
  \includegraphics[width=10cm]{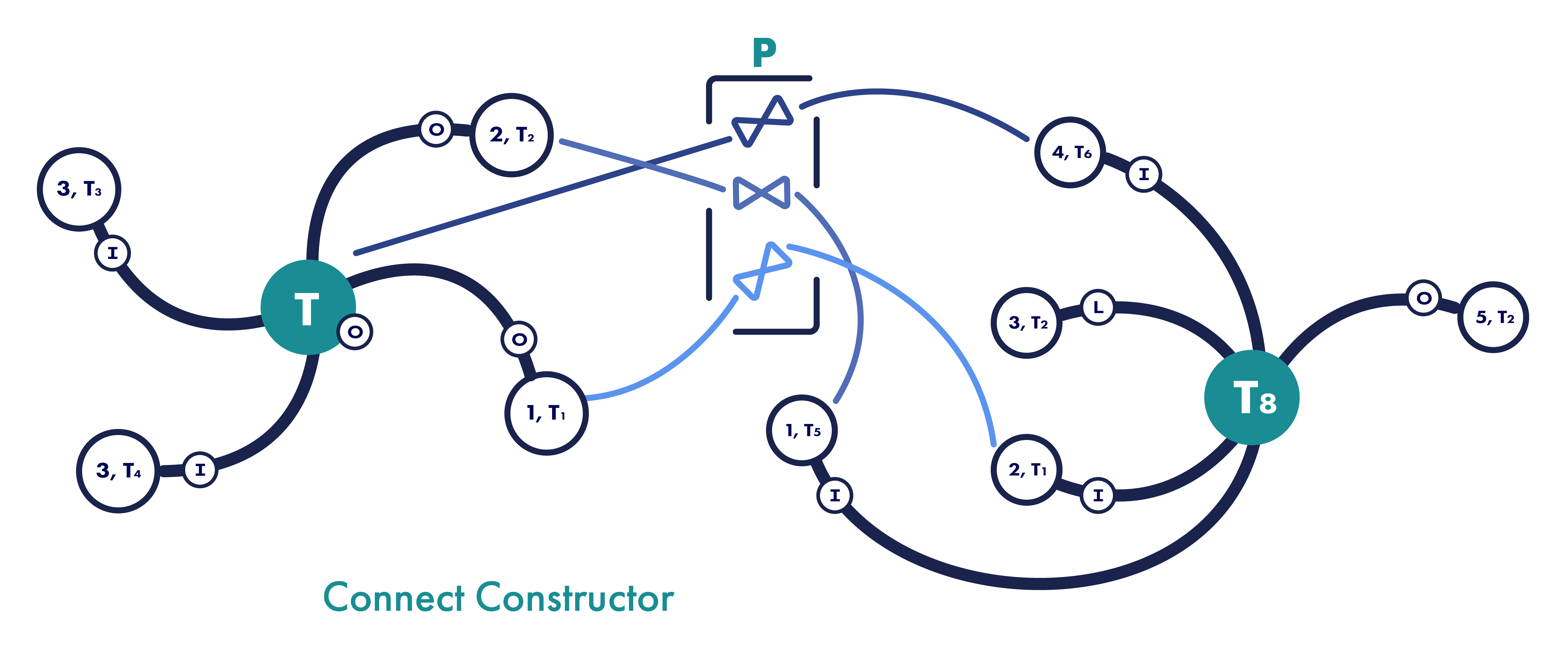}
  \caption{The Connect constructor.   The connection routing function $P$ matches outputs of one edge with inputs of another, creating a metapath (DTMG) via stringing the two together accordingly.}
  \label{fig:Connect}
\end{figure}

\subsubsection{Empty}

The empty DTMG, an edge with no targets, is the unit of both $\Vert$ and  $\bowtie_{\emptyset}$.  We also have the law

$$
G \bowtie_{\emptyset} H = G \Vert H
$$

\subsubsection{Swap}

The swap operator $\textrm{swap}_{j,k}$ maps an edge $E$ with arity $\geq max(j,k)$ into an edge $\textrm{swap}_{j,k} . E$ that is identical to $E$ except the $j$'th and $k$'th targets are swapped.   Viewed as a DTMG itself, ${swap_{j,k}}_I$ is its $j$ input arguments and ${swap_{j,k}}_O$ is its $k$ output arguments.

As with directed multigraphs we have the "swap law" that if $G$ has $n$ inputs and $p$ outputs, and $H$ has $m$ inputs and $q$ outputs, then

$$
swap_{j,k} \bowtie_I ( G \Vert H) \bowtie_I swap_{p,q} = H \Vert G
$$

\noindent where $I$ is the identity mapping.

It's easy to see that one can map any crf $P: s^k_{[m]} \rightarrow s^k_{[n]}$ of size $k$ into any other $P_1: s^k_{[m]} \rightarrow s^k_{[n]}$ of size $k$ via using some appropriate composition of multiple swap operators $\textrm{swap}_{j,k}$.

\begin{figure}
  \includegraphics[width=10cm]{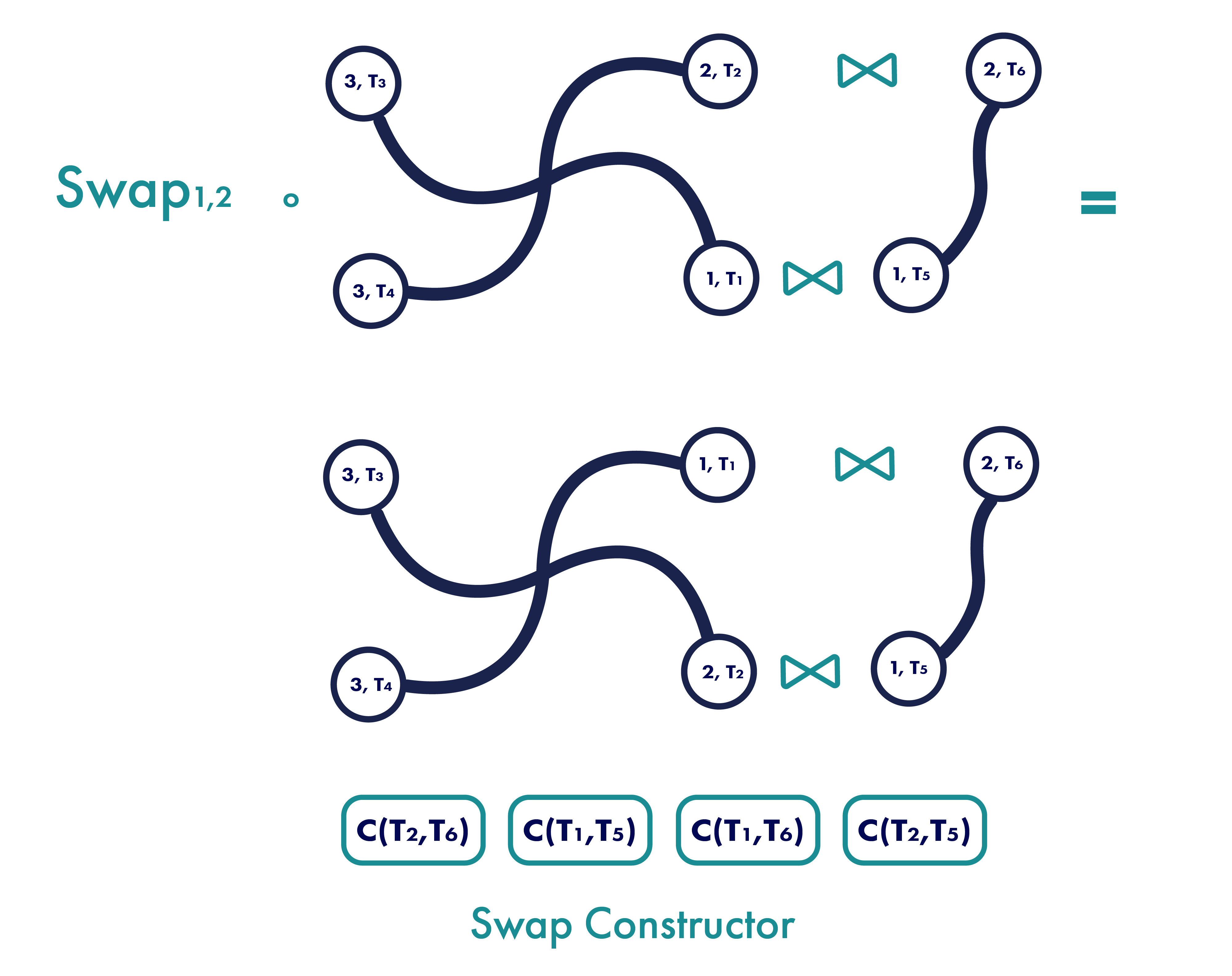}
  \caption{The Swap constructor, which acts on an edge and outputs an edge with two of its targets swapped in position.   As the figure indicates, the result of a swap can only be inserted into the same relationships as the pre-swap edge if the type compatibilities happen to work out.}
  \label{fig:Swap}
\end{figure}

\section{DTMGs as Categories}

We now outline a category-theoretic formalization of DTMGs, leveraging the construction operations defined above.

Much as Gibbons \cite{gibbons1995initial} shows the the the algebra of directed multigraphs is essentially a symmetric strict monoidal category (SSMC) enriched with objects representing vertices, similarly we show that the algebra of DTMGs is a SSMC with objects representing DTMGs and arrows representing metapaths.

A strict monoidal category (SMC) $(B, +, e)$ can be characterized in multiple ways; following Gibbons \cite{gibbons1995initial} we characterize it here as a category $B$ in which:

\begin{itemize}
\item the objects of $B$ form a monoid with respect to $+$ (as a binary operation on
objects of $B$) and $e$ (as an object of $B$)
\item the operator $+$ also acts on arrows of $B$; if $x : G \rightarrow H$ and $y : J \rightarrow K$ then
$x + y : G + J \rightarrow H+ K$ moreover, $+$ satisfies the laws
\begin{itemize}
\item $(x + y) + z = x + (y + z)$
\item $id_e + x = x$
\item $x + id_e = x$
\item $id_G + id_H= id_{G+H}$
\item $(w + x) \circ (y + z) = (w \circ y) + (x \circ z)$
\end{itemize}
provided in the last case that all the compositions are defined.
\end{itemize}

A symmetric strict monoidal category (SSMC) $(B; +; e )$ is then a SMC $(B; +; e)$
with a family of arrows $ \gamma_{G,H} : G+H \rightarrow G+H$ for all objects $G$ and $H$ of $B$, for
which the following laws hold:

\begin{eqnarray*}
\gamma_{G,0} = id_G \\
\gamma_{G,H+J} = (\gamma_{G,H} + id_J) \circ (id_H + \gamma_{G, J} ) \\
\gamma_{G,H} \circ (x + y) \circ \gamma_{J,K} = y + x \\
\end {eqnarray*}

\noindent provided in the last case that $x : H \rightarrow J$ and $y : G \rightarrow K$.

The algebra of DTMGs forms a SSMC $(B, \Vert, empty, swap)$ in which the category $B$ has as objects the DTMGs, and arrows $x : G^1 \rightarrow_P \mathcal{G^2}$ corresponding to metapaths.  Composition of arrows is carried out by the operators $\bowtie_P$, and the identity object on $G$ may e.g. be a binary edge with root type that connects $G$'s first input to $G$'s first output.

Now consider an SSMC in which the objects of the base category $B$ are DTMGs, and the collection of arrows of $B$ also contains arrows $x : G^1 \rightarrow \mathcal{G^2}$  for each pair of DTMGs $G^1, \mathcal{G^2}$. We call such a SSMC an enriched SSMC, and write $\mathcal{D}$ for the category of all enriched SSMCs (with functors between SSMCs as arrows). We define the algebra $\mathcal{M}$ of DTMGs to be the initial SSMC in the category $\mathcal{D}$. Informally, this says that DTMGs form the smallest algebra closed under the constructors in which all and only the DTMG laws hold.

The above does not consider values attached to edges, but these can be incorporated e.g. by creating additional arrows from edges to the associated values.

\section{Catamorphisms and Anamorphisms on Metagraphs}

We now construct catamorphisms (fold) and anamorphisms (unfold) on typed metagraphs, similarly to how Gibbons \cite{gibbons1995initial} defines catamorphisms on directed acyclic multigraphs.

\subsection{DTMG Catamorphisms}

A function $f: \mathcal{D} \rightarrow B$ is a DTMG catamorphism  if there exists a constant $a \in B$ and a family of constants  $e_{\mathcal L, \mathcal V} \in B$  indexed by typed-target tuples ${\mathcal L}$ and value-lists ${\mathcal V}$,  and binary operators $ \bigoplus : B \times B \rightarrow B$ and  $ \bigotimes_P : B \times B \rightarrow B$ such that

\begin{eqnarray*}
f.empty = a. \\
f.edge_{\mathcal L, \mathcal V} = e_{\mathcal L, \mathcal V} \\
f.(x  \Vert y) = f.x \bigoplus f.y \\
f.(x \bowtie_P y) = f.x \bigotimes_P  f.y \\
\end{eqnarray*}

\noindent (in fact, $f.x \bigotimes_P f.y$ need only be defined when $x$ and $y$ are compatible) and such that these constants and functions form an enriched SSMC in the obvious way.

\subsubsection{Example DTMG Catamorphisms}

Some simple examples of DTMG catamorphisms are:

\begin{itemize}
\item The function ${\it numtargets}$ which counts the number of targets in all the edges in an DTMG.
\item The function ${\it shortestpathlength}$ , which returns the length of (the number of edges in) the shortest path from one edge to another 
\item The function ${\it shortestpathlist}$ , which returns the a list containing the edges in the shortest path from one edge to another

\end{itemize}

\subsection{DTMG Anamorphisms}

A function $u: B \rightarrow \mathcal{D}$ is a DTMG anamorphism  if there exists a constant $a \in B$ and a family of constants  $e_{\mathcal L, \mathcal V} \in B$  indexed by typed-target tuples ${\mathcal L}$ and value-lists ${\mathcal V}$,  and binary operators $ \bigoplus : B \times B \rightarrow B$ and  $ \bigotimes_P : B \times B \rightarrow B$ such that

\begin{eqnarray*}
u.a = empty  \\
u.e_{\mathcal L, \mathcal V}  = edge_{\mathcal L, \mathcal V}  \\
u. (x \bigoplus y) =  u.x  \Vert u.y  \\
u. (x \bigotimes_P y) = u.x \bowtie_P u.y  \\
\end{eqnarray*}

\noindent (where $x \bigotimes y$ need only be defined when $x$ and $y$ are compatible) and such that these constants and functions form an enriched SSMC in the obvious way.

\subsection{DTMG Hylomorphisms }

A metagraph hylomorphism may then be defined as a composition $f \circ u$ where $u$ is a metagraph anamorphism and $f$ is a metagraph catamorphism.

Generally speaking, hylomorphisms are a powerful way to organize a variety of useful computations, including arbitrarily complex recursive computations.    We believe metagraph hylomorphisms can be a powerful way to organize recursive metagraph computations.

\subsection{DTMG Metamorphisms }

A metagraph metamorphism is a composition of the form $u \circ f$, which is commonly an elegant way to effect a change of representation.   For instance, if $f$ is {\it shortestpathlist}, which produces a list of the edges on the smallest metapath from given edge $E_1$ to given edge $E_2$, then $u$ would be an anamorphism that takes this list and uses it as raw material to construct another structure -- which could be for instance a list of sub-lists representing small metapaths that, concatenated, bring you from $E_1$ to $E_2$.

Gibbons \cite{gibbons2007metamorphisms} has given conditions under which metamorphisms support streaming -- incremental unfolding of the output of fold, carried out before the fold operation is complete.

\section{Memorious Metagraph Morphisms via History Hypertrees}

The metagraph morphisms described above may be tweaked in such a way as to make them retain and produce historical records of their actions as they proceed (for a general overview of this notion, outside the metagraph context, see \cite{kmett2008}).   This  yields notions of metagraph histomorphisms (memory-propagating catamorphisms), futumorphisms (memory-creating anamorphisms) and chronomorphisms (compositions of catamorphisms with futumorphisms).   These constructs embody even more flexibly pragmatically valuable ways of organizing recursive metagraph computations.

\subsection{DTMG Histomorphisms}

A histomorphism, conceptually, is a catamorphism in which, at each step of the process of folding a certain operation into a structure, the intermediate results from all the previous steps are available to the operation being folded in.

To define histomorphisms on metagraphs, we must first extend the operators $\Vert$ and $\bowtie_P$ to operate on elements of the Cartesian product $\mathcal{D} \times \mathcal{H}$, where $\mathcal{H}$ is the space of ordered forests of ternary directed hypertrees whose nodes are labeled with elements of $\mathcal{D}$ (i.e. with DTMGs) and whose links are labeled with crfs $P$.  \footnote{By an ordered forest of hypertrees is simply meant a finite ordered list of hypertrees.}   These hypertrees $H$ may be called {\it history hypertrees}.

A link in a history hypertree $H$ between nodes labeled $(G^1, G^2, G^3)$, with label $P$, indicates that $G^1$ is the DTMG resulting from performing the construction operation  $G^2 \bowtie_P G^3$.

Given this setup,

\begin{itemize}
\item $H_1 \Vert H_2$ is the concatenation of the hypertree-list $H_1$ and the hypertree-list $H_2$.
\item $H_1 \bowtie_P H_2$ is the operation that creates a history hypertree $H_3$ with a new root node $N$, such that:
\begin{itemize}
\item $N$ is labeled with the DTMG $G^3 = G^1 \bowtie_P G^2$, where $G^1$ and $G^2$ are the DTMGs labeling the root nodes of $H_1$ and $H_2$ respectively
\item The link $(G^1, G^2, G^3)$ is labeled with $P$.
 \end{itemize}
\end{itemize}

We may then say: A function $f: \mathcal{D} \rightarrow B \times \mathcal{H} $ is a DTMG histomorphism  if there exists a constant $a \in B$ and a family of constants  $e_{\mathcal L, \mathcal V} \in B$  indexed by typed-target tuples ${\mathcal L}$ and value-lists ${\mathcal V}$,  and binary operators $ \bigoplus : (B \times \mathcal{H}) \times (B \times \mathcal{H} )\rightarrow B \times \mathcal{H}$ and  $ \times_P : (B \times \mathcal{H})  \times (B \times \mathcal{H}) \rightarrow B \times \mathcal{H}$ such that

\begin{eqnarray*}
f.empty = a \times a \\
f.edge_{\mathcal L, \mathcal V} = e_{\mathcal L, \mathcal V} \times e_{\mathcal L, \mathcal V} \\
f.(x \times H_x  \Vert y \times H_y) = f.x \times H_x \bigoplus f.y \times H_y \\
f.(x \times H_x  \bowtie_P y \times H_y) = f.x \times H_x \bigotimes_P  f.y \times H_y \\
\end{eqnarray*}

\noindent (where $f.x \times H_x \bigotimes_P f.y \times H_y$ need only be defined when $x$ and $y$ are compatible) and such that these constants and functions form an enriched SSMC in the obvious way.

\subsection{DTMG Futumorphisms}

A futumorphism, conceptually, is the opposite of a histomorphism: it's an anamorphism with memory.   In a futumorphism,  at each step of the process of building a structure via unfolding, the intermediate results from all the previous steps are available to the operation doing the building.

To define futomorphisms on metagraphs, we leverage the same formalism of history hypertrees used for histomorphisms.

We may say that: A function $u: \mathcal{D} \rightarrow B \times \mathcal{H} $ is a DTMG futumorphism  if there exists a constant $a \in B$ and a family of constants  $e_{\mathcal L, \mathcal V} \in B$  indexed by typed-target tuples ${\mathcal L}$ and value-lists ${\mathcal V}$,   and binary operators $ \bigoplus : (B \times \mathcal{H}) \times (B \times \mathcal{H}) \rightarrow B \times \mathcal{H}$ and  $ \bigotimes_P : (B \times \mathcal{H})  \times (B \times \mathcal{H}) \rightarrow B \times \mathcal{H}$ such that

\begin{eqnarray*}
u.( a \times a) = empty\\
u.( e_{\mathcal L, \mathcal V} \times e_{\mathcal L, \mathcal V}) = u.edge_{\mathcal L, \mathcal V}   \\
u. ( x \times H_x \bigoplus y \times H_y) = u.(x \times H_x ) \Vert u.(y \times H_y)  \\
u. (x \times H_x \bigotimes_P  y \times H_y) = u.(x \times H_x)  \bowtie_P u.(y \times H_y)  \\
\end{eqnarray*}

\noindent (where $u. (x \times H_x \bigotimes_P  y \times H_y)$ need only be defined when $x$ and $y$ are compatible) and such that these constants and functions form an enriched SSMC in the obvious way.

\subsection{DTMG Chronomorphisms}

A chronomorphism is basically a "hylomorphism on history" -- i.e. a composition of an unfolding (anamorphism) operation with a (re-)folding (catamorphism) operation.

A metagraph chronomorphism can be defined as a composition $f \circ u$ where $u$ is a metagraph futumorphism and $f$ is a metagraph histomorphism.

Unlike simpler hylomorphisms, chronomorphisms have not yet been used extensively in the functional programming world.    However, we believe that metagraph chronomorphisms can be a powerful framework for formulating diverse AI algorithms in a common framework, which is the major motivating factor behind this paper.

The analogue of a metamorphism, but using history, might be called a "metachronomorphism" \footnote{This may have another name in the literature, which the author has so far failed to uncover}.   Here one does a histomorphism first and then a futumorphism on the results, so $u \circ f$.   The futumorphism is able to use the history hypertree accumulated while executing the histomorphism to help guide its operations.

It is also possible to combine these, and nest a metachronomorphism within a chronomorphism, in the pattern $f \circ u_1 \circ f_2 \circ u$ -- this e.g. may mean unfolding, then doing a change of representation, then refolding.

\section{Connector-Ordered DTMGs}

In this section we briefly mention an alternate way of setting up the constructs that we've been exploring.   At the moment this alternative seems overall not preferable to us, but it does have certain advantages and we feel it may have some uses in future.

The parameter $P$ in the connection-operation $G \bowtie_P H$ is sensible and simple enough to deal with in practice, but also a complicating factor.  An alternative approach is to wrap the operator $P$ into the source  metagraph $G$.

By a Connector-ordered Directed Typed Metagraph (ConDTMG), we will mean an ordered pair $(G,P)$ where $P$ is a crf that is compatible with channeling output out of $G$.

The Beside, Empty and Swap constructors given above work without change for ConDTMGs.  An additional ConnectionRoutingFunction constructor that builds a crf $P$ is required, along with an operation that glues a crf to a DTMG to get a ConDTMG pair.   The Connect constructor is then modified to  take the forms

\begin{eqnarray}
(G,P) \bowtie H \\
(G,P) \bowtie (H, P_1)  
\end{eqnarray}

\noindent where in each case it only works if $P$ can successfully route $G$ into $H$.

In the ConDTMG case, the composition of arrows (metapaths) is carried out by the $\bowtie$ operator without need for parametrization.

The various morphisms work out similarly to with ordinary DTMGs, but with different equations for the connection operator.

For catamorphisms: 

$$
f.(x \bowtie y) = f.x \bigotimes  f.y \\
$$

For anamorphisms:

$$
u. (x \bigotimes y)  = u.x \bowtie u.y \\
$$

For histomorphisms:

$$
f.(x \times H_x \bowtie y \times H_y) = f.x \times H_x \bigotimes  f.y  \times H_y  \\
$$

For futumorphisms:

$$
f.(x \times H_x  \bowtie y \times H_y) = u. (x \times H_x) \bigotimes  u. (y \times H_y) \\
$$

In the following section we will go back to DTMGs for simplicity, however the basic concepts also apply to ConDTMGs.

\section{Metapaths and Morphisms on Undirected Typed Metagraphs}

We have dealt so far mainly with DTMGs, but it's actually possible to replicate what we've done above for undirected typed metagraphs as well.   For this case we must replace the Connect constructor above with a different Connect$_Q$ constructor, defined as follows.  Given two edges $E_1$ and $E_2$ and a qcrf ("quadratic connection routing function") $Q$, the Connect$_Q$ constructor $\bowtie_Q$ creates the connections between $E_1$'s targets and $E_2$'s targets that $Q$ specifies.

The basic idea is that the Connect$_Q$ constructor here has the potential to connect every one of $E_1$'s targets to every one of $E_2$'s targets, however there are many ways to group such connections into edges.   A given target $t$ of $E_1$ may be matched directly to one target of $E_2$, or it may be assigned as a target of a newly created edge whose other targets are drawn from those of $E_1$ and $E_2$.   In the case of newly created edges, the types of the new edge targets should be precisely matched to the types of the connected $E_1$ and $E_2$ targets.   The function $Q$ needs to specify how all this mapping is to be done.

This is a generalization of the Connect$_Q$ constructor used to generate ordinary unlabeled or labeled graphs in \cite{mokhov2017algebraic}.   In those simpler cases, the Connect$_Q$ constructor between subgraph $S_1$ and subgraph $S_2$ simply connects every vertex of $S_1$ with every vertex $S_2$, using a newly created link.   This is obviously a sub-case of the constructor $\bowtie_Q$ specified here.

\begin{figure}
  \includegraphics[width=10cm]{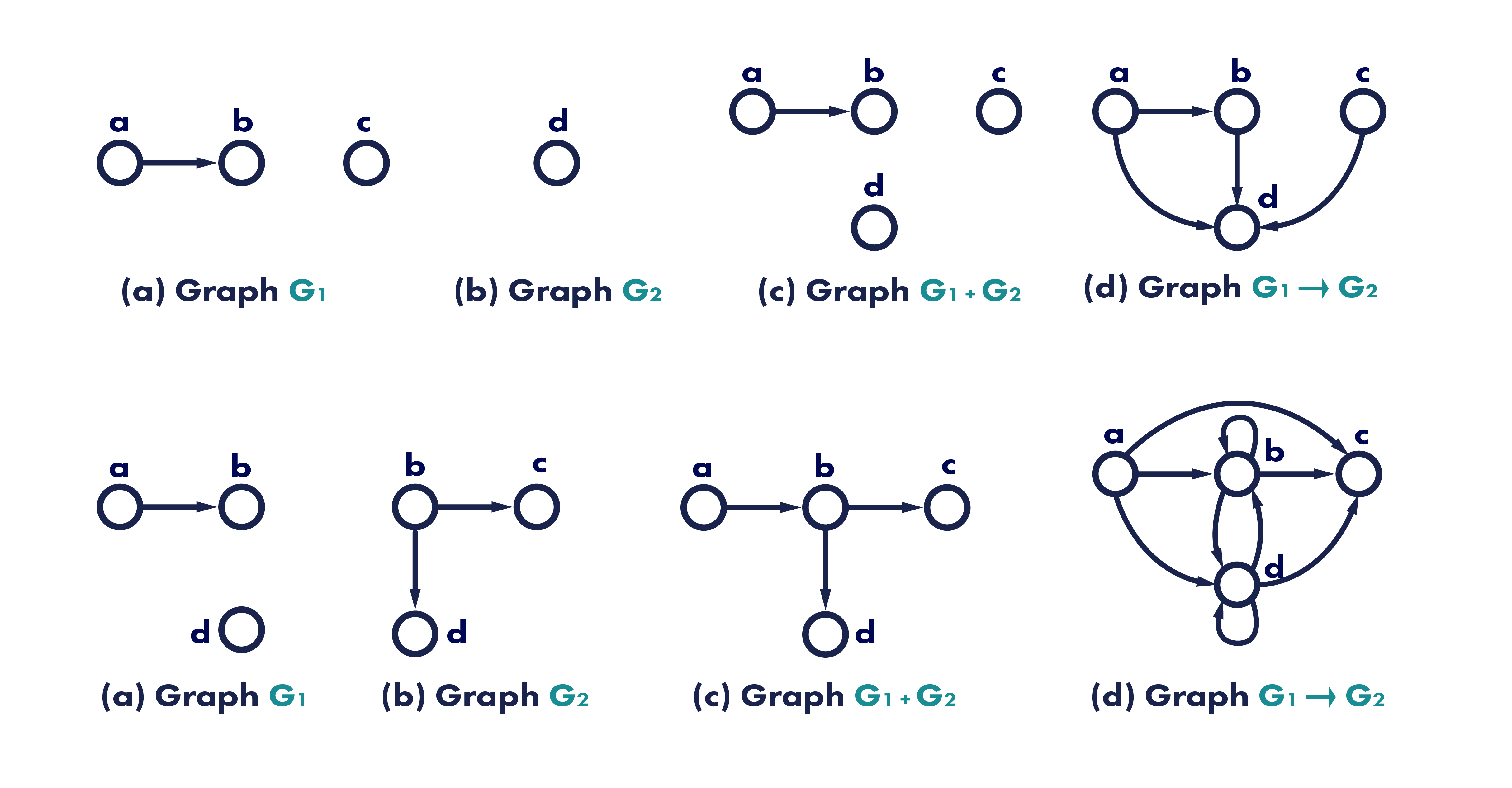}
  \caption{The overlay (top) and Connect (bottom) constructors on labeled graphs from \cite{mokhov2014algebra}, of which the Union and Connect$_Q$ constructors described for undirected metagraphs is a generalization.}
  \label{fig:Mokhov}
\end{figure}

Given 

\begin{itemize}
\item the Connect$_Q$ constructor
\item a Union constructor that simply takes the set-theoretic union of two edge-sets (which may or may not overlap)
\item an edge constructor that builds a new edge
\item an empty constructor
\end{itemize}

\noindent then, following the arguments of \cite{mokhov2017algebraic} and \cite{mokhov2014algebra}, we can see that we have process able to construct all TMGs without requiring any division of the targets of the edges into inputs and outputs.   One arrives at a notion of a TMG (not DTMG) metapath as a sequence of edges $E_1, E_2, \ldots, E_n$ where each edge is connected to its neighbors in the sequence.   

Using this approach, the various DTMG morphisms discussed above end up working similarly for TMGs.  Whether it's more appropriate to look at directed or undirected metapaths depends on the application domain.

\section{Metagraph Processes and Forested TMGs}

One can also extend these various morphisms to deal with the multiple DTMGs that may be defined on a single undirected TMG, a step which may be useful (alongside other possible applications) for modeling processes acting on metagraphs.

Given a typed metagraph, and a process that incrementally traverses through the TMG connecting edge-targets to other edge-targets, one can then define a DTMG where the temporal order of the traversal defines the direction of the paths through the DTMG.   (We'll go on to processes that update values associated with edges in a moment, but for now let's look at processes that just perform connections between edges.)

Suppose such a process traverses from $E_1$ to $E_2$, in a way that binds the targets of $E_1$ with indexes in ${\mathcal l}_1 = \{ t_{11}, \ldots , t_{1r_1}  \}$ to the targets of $E_2$ with indexes in ${\mathcal l}_2 = \{ t_{21}, \ldots , t_{2r_2}  \}$.   Then we can consider $E_1$ as having output-target-set corresponding to ${\mathcal l}_1 = \{ t_{11}, \ldots , t_{1r_1}  \}$ and $E_2$ as having input-target-set corresponding to ${\mathcal l}_2 = \{ t_{21}, \ldots , t_{2r_2}  \}$, and consider there to be a metapath from $E_1$ to $E_2$.    

A similar analysis holds if one has a process that traverses from a larger multigraph (bigger than a single edge) to another multigraph.

A collection of multiple processes acting on a typed metagraph, or a process acting on the same typed metagraph over time in a way that allows it to repeatedly visit the same edges, will then in general create a forest of DTMGs living on the same TMG.   Different DTMGs in the same forest may involve the same edge but broken down into input, output and lateral target-sets in different ways.   We may refer to this sort of construct as an FTMG, a Forested Typed MetaGraph.

The various morphisms described in the previous sections can be extended straightforwardly to FTMGs via considering a forest of DTMGs as a list of DTMGs.   Gibbons \cite{gibbons1995initial} gives a (basically trivial) initial algebra formulation for lists, along the same lines as his treatment of directed multigraphs.   So by composing our metagraph morphisms with corresponding list morphisms, one obtains the full menagerie of morphisms on FTMGs.   

The extension to the case where the different DTMGs in the forest are weighted with probabilities is also straightforward -- then in place of the list one represents the forest as a probability distribution.   But folding over this distribution is essentially just folding over a one-dimensional linkage graph with a weight at each node, a case covered by  \cite{gibbons1995initial} in the guise of labeled linkage graphs.    As one potentially interesting application of this, if the underlying types on edges and targets are probabilistic, this will naturally lead to probability weightings on the DTMGs in the forest.

\subsection{Encompassing More Realistic Metagraph Processes}

With FTMGsaspects of the ways metagraphs are used in applications such as the OpenCog system.   To get all the way there, though, it will be valuable to address a few additional aspects of the metagraph-manipulation processes that are commonly seen in practice.

First, processes that create new metagraph edges can be treated in basically the same manner outlined above.   One can consider a huge {\it virtual TMG} consisting of all possible edges up to some large finite size $N$, and then a smaller {\it realized TMG} drawn from the virtual TMG.   A process acting on the realized TMG, which would normally be thought of as creating new edges (of size less than $N$) that connect to one or more existing edges in the realized TMG, can instead be conceived as expanding the realized TMG via pulling new edges from the virtual TMG into it.   

In the 2020 OpenCog version, for instance, the essential Pattern Matcher tool  \cite{VepstasPatternMatcher2020} operates via QueryLinks, which are logically decomposable into MeetLinks and PutLinks \cite{VepstasQueryLink2020}.  A MeetLink connects a metagraph representing an abstract query involving variables with matching metagraphs that fit into the slots presented by these variables.   A PutLink creates a new metagraph formed by taking the slot-fillers from the matching metagraphs and actually inserting them in the slots of the query metagraph.   In the present view, the connections between the metagraph query posed to MeetLink and the metagraphs fitting into its slots form metapaths, which are (when the MeetLink is created as part of a pattern matching dynamic) parts of a larger metapath comprising the Pattern Matching process and results.  The connections between the metagraph query posed to the PutLink and the new metagraph created by the PutLink are also metapaths forming parts of this larger metapath through the virtual TMG.

Processes that update values associated with edges, based in part on values associated with other edges, can be fit into the same formal framework by considering each value-update as the creation of a new edge, and a replacement of connections to the old edge with connections to the new one.  That is, if we start with a base TMG that has mutable values associated with its edges, then we create a virtual TMG that contains versions of each edge in the base TMG with all possible value-assignments to its mutable values, drawn from some (potentially large) finite range.    Value-update processes can then be represented as processes that create new connections to edges, often including edges that were not previously in the realized TMG.

Finally, to conveniently represent "forgetting" -- removal of an edge from the realized TMG -- in this framework, one approach is to create an integer value signifying "forgotten."    Suppose an edge is denoted by a pair $(\mathcal L, \mathcal V_{\neg F})$, where $\mathcal V_{\neg F}$ comprises all the values associated with the edge except the {\it forgotten} value.   Then each time the edge is removed from or reinserted into the realized TMG, the {\it forgotten} value may be incremented by $1$.   One may then define the {\it pruned TMG} as follows: It is mostly the same as the realized TMG, but if the largest {\it forgotten} value that associated with the edge $(\mathcal L, \mathcal V_{\neg F})$ in any instance $(\mathcal L, \mathcal V)$i n the realized TMG is odd, the edge should not be included in the pruned TMG even if it's in the realized TMG.

FTMGs on the realized TMG appear to be a general and robust approach to representing a variety of practical dynamics on metagraphs.   This provides a reasonably elegant framework within which to explore the formulation of useful metagraph algorithms and dynamics in terms of the various species of morphisms described above -- an important and intriguing task which however goes beyond the bounds of this paper.

Finally, it's worth emphasizing that the practical software implementation of metagraphs and processes thereupon will involve numerous considerations not addressed here, and the most efficient and effective software structures may not correspond in the most direct way to the underlying mathematics or concepts.    Our objective here has merely been to outline a formal model that captures the key aspects of metagraphs and (many of the) algorithms acting upon them in a generic and relatively simple way.

\section{Topology and Algebra on Metagraphs}

Finally we review some further mathematical structures definable on metagraphs, which we have found useful for work applying metagraphs and their morphisms in an AI context.   We use metapaths to define a topology on metagraphs; this topology leads us to a Heyting algebra structure on metapaths, which then brings us to the verge of metagraph-based intuitionistic logic and intuitionistic probabilities (the latter topics however will be left for a subsequent paper).

\subsection{Metagraph Homomorphisms}

The notion of graph homomorphism can be extended naturally to TMGs, though the bookkeeping becomes a little more complicated due to the types.

In an ordinary graph, each elementary homomorphism is created by identifying two non-adjacent nodes.  One can then show that every graph homomorphism is a composition of elementary homomorphisms.

In a typed metagraph $M$, an elementary homomorphism may be defined as an identification of two subgraphs $S_1$ and $S_2$ of $M$, where the mapping between the two subgraphs is lossless regarding the targets and types of the subgraphs.    That is, suppose there is a mapping from $S_1$ to $S_2$ where the targets of $S_1$ are mapped $1-1$ onto the  targets of $S_2$ in such a way that 

\begin{itemize}
\item if some of the targets of edge $E_1$ are mapped into some of the targets of edge $E_2$, then $E_1$ and $E_2$ have the same type 
\item  targets mapped into each other have equivalent types
\item if two targets of $S_1$ have the same index number on the same edge, they are mapped into two targets of $S_2$ that are on the same edge and have the same index number; and vice versa
\end{itemize}

\noindent Then we can create a new metagraph $M'$ in which every target connecting to some target of $S_1$, is instead connected to the corresponding target of $S_2$ according to the mapping.

It is evident that every metagraph homomorphism can be expressed as a composition of elementary metagraph homomorphisms.

\subsection{The Metapath Topology}

To think about continuity or "smoothness" of metagraph transformations, one needs to define a topology on the space of subgraphs of a metagraph, i.e. one needs to identify a collection of subgraphs suitable to serve as the open sets of a topology.   

One way to define a topology for DTMGs is to follow the topology for digraph edges from \cite{abdu2018topologies} .  What they do is to take the collection of length-2 paths in a digraph as the subbasis for generating a topology on the digraph.   Basically this yields a topology in which the open sets are sets of paths through the digraph.  

Similarly, we can create a topology on DTMGs in which the open sets are the metapaths, by starting with all metapaths of length 2 as a subbasis.   As a set of metapaths is itself a metapath, we can deviate from the digraph case by just stating that metapaths are the open sets (rather than bothering to look at sets of metapaths).   We must also add the empty set and the whole metagraph (metapaths to and from nowhere) as open sets.  

The metagraph topology can also be defined on a forested TMG, by taking the open sets on a forested TMG $M$ as pairs $(i,P)$ where $P$ is a metapath in the $i$'th DTMG in the forest.

That metapaths meet the criteria to serve as the collection of open sets should be clear.   The union or intersection of two metapaths is a metapath.

To define smooth transformations on metagraphs in a way that matches with metapath topology, one can modify the definition of metagraph homomorphism given above to make it asymmetric. 

Define an {\it elementary smooth transformation} from subgraph $S_1$ to subgraph $S_2$ of $M$ as a mapping from $S_1$ to $S_2$ where the targets of $S_1$ are mapped (not necessarily $1-1$) onto the  targets of $S_2$ in such a way that 

\begin{itemize}
\item if some of the targets of edge $E_1$ are mapped into some of the targets of edge $E_2$, then $E_2$'s type inherits from that of $E_2$ 
\item  if a target $t_1$ in $S_1$ is mapped into a target $t_2$ in $S_2$, then the type of $t_2$ inherits from the type of $t_1$
\item if two targets of $S_1$ have the same index number on the same edge, they map into targets of $S_2$ that are on the same edge and have the same index number (but not necessarily vice versa)
\end{itemize}

\noindent What this definition guarantees is basically that any edge that connects to $S_2$, will also be able to connect to $S_1$.   

As the name suggests, the intent here is that an elementary smooth transformation on a metagraph should be continuous in a topological sense.    And this does seem to hold up in the metapath topology.

In the metapath topology, it is clear that if one has a metapath $P_2$ passing through $S_2$, and an elementary smooth transformation $F$ mapping $S_1$ into $S_2$,  then $F^{-1}(P_2)$  will be a metapath passing through $S_1$ (or else the empty set, which is also an open set in the topology).   

It is also evident that any continuous mapping from a DTMG $\mathcal{D}$ into itself can be represented as a composition of elementary smooth transformations.

This gives us a notion of what it means for one metapath to be "continuously deformed" into another, which allows us to define homotopy groups on metapaths -- equivalence classes of metapaths that are continuously intertransformable, even if not homomorphic.   When considering representation of logical proofs as metapaths, this leads us directly to a metagraph instantiation of homotopy type theory (a topic beyond the scope of this paper but to be touched in intended sequels).

\subsubsection{Continuity of Metagraph-to-Metagraph Morphisms}

It is also interesting to consider continuity of the menagerie of metagraph morphisms considered above, especially in the case where $\mathcal{B} = \mathcal{D_1}$, i.e. the target of a catamorphism or the source of an anamorphism is itself a metagraph.   Let us call these M2M morphisms, i.e. Metagraph-to-Metagraph.   

In the case $\mathcal{D} = \mathcal{D_1}$ one has folding and unfolding processes that are operating within the confines of a single metagraph, and are concerned with relating different operations ($\bigoplus$ and $\bigotimes$) on the metagraph to the $\Vert$ and $\bowtie$ constructor operations.

It's easy to see that in the case of M2M morphisms, we have

\begin{thm}

\begin{itemize}
For any DTMG,
\item  M2M anamorphisms and futumorphisms are continuous in the metapath topology
\item  If the operators $\bigoplus$ and $\bigotimes$ are continuous in the metapath topology, then M2M catamorphisms, histomorphisms, hylomorphisms and chronomorphisms are continuous in the metapath topology 
\end{itemize}
\end{thm}

\subsection{Metapath Algebra}

Every topology induces a Heyting algebra, and the Heyting algebra induced by the metapath topology on a DTMG is an interesting and relevant one.   

We can define a partial order $\leq$ on metapaths via: $A \leq B$ if $A$ is a subpath of $B$.

Meet and join operations are then defined in the obvious way, simply as union and intersection.   

An implication $A \rightarrow B$ is 
defined as $~A \cup B$, i.e. as the union of the metapath $B$ with the largest metapath in $\mathcal{D} - A$.

The relative pseudo-complement $A \rightarrow B$ is defined as the largest $x$ so that $A \cap x \leq B$.   So this is the metapath comprised of $B$ unioned with the biggest metapath that is in $\mathcal{D}-A$.    

The pseudo-complement $~A$ of a metapath $A$ in DTMG $\mathcal{D}$ is defined as  $A \rightarrow 0$, which is the largest metapath in the set-theoretic complement $\mathcal{D} - A$.   Noteworthy is that this does not obey the rule $~A \lor A = \mathcal{D}$.   

Every Heyting algebra induces an intuitionistic logic \cite{heyting1930}, and so this construction allows us to build logics on metagraphs in a natural way.   One can also define intuitionistic probabilities using Heyting algebras in place of Boolean algebras \cite{Goertzel17a}, so we also have here a path to defining probabilities that are natural to metagraphs.   These directions bring us beyond the scope of this paper, but are highly relevant to our reasons for pursing metagraph morphisms in the first place.

\section{Conclusion and Future Directions}

We have shown that, in the context of quite general metagraphs, catamorphisms, anamorphisms and more refined similar morphisms take fairly simple and elegant forms.    It appears these concepts can be applied to realistic metagraphs involving complexities like dependent and probabilistic types, multidimensional values and dynamic processing including insertion and deletion of edges.   The next natural step in this research programme would be to formulate various useful metagraph-based algorithms and algorithm classes in terms of these morphisms.

\bibliographystyle{alpha}
\bibliography{bbm}

\end{document}